# Attributed Network Embedding Model for Exposing COVID-19 Spread Trajectory Archetypes

Junwei Ma[1*], Bo Li[2], Qingchun Li[3], Chao Fan[4] and Ali Mostafavi[5]

[1] (Corresponding author) Ph.D. Student, Urban Resilience.AI Lab, Zachry Department of Civil and Environmental Engineering, Texas A&M University, College Station, Texas, United States; e-mail: jwma@tamu.edu

[2] Ph.D. Student, Urban Resilience.AI Lab, Zachry Department of Civil and Environmental Engineering, Texas A&M University, College Station, Texas, United States; e-mail: libo@tamu.edu

[3] Post-doctoral Research Associate, Urban Nexus Lab, Department of Civil and Environmental Engineering, Princeton University, Princeton, New Jersey, United States; e-mail: qingchun.li@princeton.edu

[4] Assistant Professor, School of Civil and Environmental Engineering and Earth Sciences, Clemson University, Clemson, South Carolina, United States; e-mail: cfan@g.clemson.edu

[5] Associate Professor, Urban Resilience.AI Lab Zachry Department of Civil and Environmental Engineering, Texas A&M University, College Station, Texas, United States; e-mail: amostafavi@civil.tamu.edu

**Abstract:** The spread of COVID-19 revealed that transmission risk patterns are not homogenous across different cities and communities, and various heterogeneous features can influence the spread trajectories. Hence, for predictive pandemic monitoring, it is essential to explore latent heterogeneous features in cities and communities that distinguish their specific pandemic spread trajectories. To this end, this study creates a network embedding model capturing cross-county visitation networks, as well as heterogeneous features related to population activities, human mobility, socio-demographic features, disease attribute, and social interaction to uncover clusters of counties in the United States based on their pandemic spread transmission trajectories. We collected and computed location intelligence features from 2,787 counties from March 3 to June 29, 2020 (initial wave). Second, we constructed a human visitation network, which incorporated county features as node attributes, and visits between counties as network edges. Our attributed network embeddings approach integrates both typological characteristics of the cross-county visitation network, as well as heterogeneous features. We conducted clustering analysis on the attributed network embeddings to reveal four archetypes of spread risk trajectories corresponding to four clusters of counties. Subsequently, we identified four features—population density, GDP, minority status, and POI visits—as important features underlying the distinctive transmission risk patterns among the archetypes. The attributed network embedding approach and the findings identify and explain the non-homogenous pandemic risk trajectories across counties for predictive pandemic monitoring. The study also contributes to data-driven and deep learning-based approaches for pandemic analytics to complement the standard epidemiological models for policy analysis in pandemics.

# 1. Introduction

COVID-19 exposed the complexity of pandemic transmission trajectories (Castro, Kim et al. 2021). One aspect of such complexity is the heterogeneity of spread trajectories in different cities and communities (Li, Yang et al. 2021). The heterogeneity of pandemic spread risk is rooted in differences in attributes of cities and communities in terms of urban characteristics, population activities and mobility, socio-demographics, and social interactions (Benzell, Collis et al. 2020, Dowd, Andriano et al. 2020, Jia, Lu et al. 2020, Ramchandani, Fan et al. 2020). Hence, in order to predictively monitor pandemic spread trajectories in devising effective non-pharmaceutical policies, it is essential to understand and distinguish cities and communities based on their pandemic transmission trajectories and the underlying influencing features. Such capability cannot be achieved using the existing standard epidemiological models (Jewell, Lewnard et al. 2020); however, machine learning-based techniques (Li, Yang et al. 2021) can provide insight into transmission trajectories. In particular, network analytic techniques have shown great potential in revealing spatio-temporal dynamics of the pandemic due to their capability of capturing spatial interactions as well as heterogeneous features of spatial areas (Geng, Gao et al. 2022).

To this end, in this study, we used attributed network embedding that captures spatial interactions among U.S. counties, as well as counties' heterogeneous features related to population activities, human mobility, socio-demographic attributes, disease attributes, and social interaction features in classifying counties based on pandemic transmission risks. We used various datasets related to the first wave of the pandemic in the United States (March–June 2022) in creating and testing the model. The attributed network embedding technique captured both typological characteristics of a cross-county movement network, as well as county-level features related to population activities, human mobility, socio-demographics, disease attributes, and social interactions. Then the counties were clustered and grouped weekly based on the results of attributed network embeddings. The stable patterns presented by each archetype reveal the heterogeneity of the pandemic spread risk across clusters of counties.

The paper unfolds as follows: we first explain the county-level features considered in this study based on the review of the extant literature related to factors influencing COVID-19 spread risks. Then a description of the dataset and methods is provided. The subsequent sections present the results of the analysis and major findings related to clusters of counties and the underlying features that distinguish transmission risk trajectories across the clusters of counties.

## 2. Features influencing the spread of COVID-19

Various features influence the transmission risks of COVID-19 pandemic in a city or county. The basic reproduction number, $R_0$, is a fundamental metric that gauges the number of new infections caused by a single infected individual in a fully susceptible population and in the absence of interventions (Diekmann, Heesterbeek et al. 1990, Anderson and May 1992). $R_0$ has been used as an important parameter to assess the potential for disease invasion and persistence in many studies (e.g. (Mohd and Sulayman 2020, Pedersen and Meneghini 2020)). However, as Shaw and Kennedy (2021) pointed out, the reproductive number alone neither could explain future dynamics of the epidemic, nor proved predictive enough to estimate the scale of an epidemic. This argument inspired researchers to consider other factors and features when assessing the spread risk of COVID-19.

Apart from disease-related features, socioeconomic features became one of the main research foci because of the disproportionate number of confirmed cases across different sub-populations (Liu, Liu et al. 2021). For example, Maiti, Zhang et al. (2021) confirmed the strong positive association between the factors such as crime and income with the cases of COVID-19. Kashem, Baker et al. (2021) highlighted an influential role of social vulnerability, which is reflected by household characteristics and race/ethnicity, in COVID-19 prevalence. Mansour, Al Kindi et al. (2021) found population aged 65 and greater and health variables were statistically significant determinants of COVID-19 incident rates and varied geographically. Saadat, Rawtani et al. (2020) found that members of households of larger size would have greater chance of infection because a larger number of family members means a larger contact network.

In addition to disease-related features and socio-demographics, pandemic spread is influenced by population activities and social interactions. Several features, including social distancing (Aquino, Silveira et al. 2020, Badr, Du et al. 2020, Qian and Jiang 2020), visits to points of interest (Zhang, Darzi et al. 2020, Chang, Pierson et al. 2021, Yuan, Liu et al. 2022), trip distance (Gao, Fan et al. 2021), and interpersonal contact density (Dargin, Li et al. 2021, Verma, Yabe et al. 2021) have been shown to have positive relationships with the COVID-19 spread risk.

In addition to local population activities, researchers also have pointed out that long-distance population movement has driven the spread of COVID-19 during the initial wave (Alessandretti 2022). There have been several studies examining the effect of human mobility across regions on the transmission risk of COVID-19 from various spatial scales. For example, Murano, Ueno et al. (2021) examined the impact of restricted domestic travel via public transportation network on transmission of COVID-19 infection; Lai, Ruktanonchai et al. (2021) used population movement data derived from mobile phones to measure the intensity and timing of global travel, and built a transmission model to simulate COVID-19 spread; Fan, Lee et al. (2021) utilized Facebook cross-county population co-location data to examine the relationship between population co-location and travel reduction and the spatio-temporal transmission risk of COVID-19 in the United States. Hence, in addition to county-level features, it is also important to capture the cross-county movement networks in examining trajectories of COVID-19 across clusters of counties at the national level.

Actually, the cross-county movement network can be constructed as a spatial network, with counties connected by population flow between them. The structural characteristics of the mobility network also affect the spread risk. For example, areas which have more connections with other areas via population movement are more likely to suffer from higher infection rates. Network topology, which describes the way nodes connect with each other, is one of the most commonly used metrics to capture network structure. The topological characteristics could be mapped by a low-dimension vector representation, which is called network embedding (Tang, Qu et al. 2015, Wang, Cui et al. 2016). Network embeddings can well preserve network proximity, which could benefit various network analysis task, such as node classification, link prediction, and network clustering (Narayanan, Belkin et al. 2006, Zhou, Huang et al. 2006, Von Luxburg 2007, Tang, Aggarwal et al. 2016). For spatial networks, network embeddings incorporate spatial dependence to boost house price prediction (Das, Ali et al. 2021) and extract structural information from the road network (Jepsen, Jensen et al. 2018). In a similar way, network embedding could be potentially helpful to capture the structural information of the spatial network constructed by cross-county movement networks.

While the effects of individual county-level features and cross-county movements on COVID-19 spread risk has been studied separately, only a limited number of studies have harnessed the capability of graph deep-learning models to capture the intertwined county-level features and cross-county spatial networks simultaneously in examining pandemic spread risk. For example, Ramchandani, Fan et al. (2020) proposed a deep-learning based DeepCOVIDNet model to forecast the range of increase in COVID-19 infected cases in future days and to compute equidimensional representations of multivariate time series and multivariate spatial time-series data. The proposed model can take in a large number of heterogeneous county-level features and learn complex interactions between these features. However, one limitation for this study is that the proposed model cannot be well interpreted due to its complexity. In other words, there is currently no suitable method to determine which exact census tract features interact with which exact disease-related features, which provides the opportunity for this study.

The transmission trajectory patterns of COVID-19 in communities are complex and are affected by interactions among various heterogeneous features. At the county scale, features related to disease attributes, socio-economic characteristics, and population activities are closely intertwined and strongly interact with each other, which results in a complicated and non-linear influence towards the spread pattern of COVID-19. At the national scale, cross-county population movements connect different areas and shape them into a spatial network, which brings into focus the complex network-related effect of the pandemic. Taking these non-linear interactions into consideration, we created an attributed network embedding model based on a cross-county movement network across U.S. counties and examined several heterogeneous county-level attributes to classify U.S. counties based on their pandemic transmission risks and also to reveal the important features that shape the distinct trajectories in each county archetype.

**3. Data and Features**

**3.1 Study area**

In this study, from March 3 through June 29, 2020, we collected features of 2787 counties in the United States. In the United States, the first confirmed COVID-19 case occurred on January 19, 2020, in Snohomish County, Washington, followed by the rapid spread of the virus across the country. The United States became the new epicenter of the disease as it surpassed Italy in terms of confirmed cases on March 24, 2020 (WHO 2020). As of June 29, 2020, there had been a total of 2,268,753 confirmed cases (25.2% of global cases) and 119,761 deaths (25.5% of global deaths) in the United States alone (University 2020). To decrease the contact and thus the transmission rate of COVID-19, many states implemented state- or local-level stay-at-home policies as well as the closure of non-essential services starting mid-March 2020. The period of analysis in this study focuses on the first wave and initial outbreak of the pandemic. Different studies highlighted the significance of this initial period for disrupting pandemic trajectories (Fan, Lee et al. 2021, Li, Yang et al. 2021). Data-driven models can be impactful in predictive pandemic monitoring and policy formulation during this stage of pandemics. Thus, our study focused on the initial outbreak stage of the pandemic.

**3.2 Datasets**

In this study, we examined 15 features related to social demographics, population activities, human mobility, social interaction, and disease attributes. The 15 features serve as a basis for dividing counties into distinct archetypes and for inferring COVID-19 spread risks. Table 1 shows the characteristics and sources of these features.

Table 1. Collected features for the data-driven model.

| Datasets | Features | Characteristics | Sources |
|---|---|---|---|
| Social Demographic | Population Density (PD) | Constant feature | U.S. Centers for Disease Control and Prevention |
| | Gross Domestic Product (GDP) | Constant feature | U.S. Department of Commerce |
| | Overall COVID-19 Community Vulnerability Index (CCVI)<br>• Socioeconomic status<br>• Household composition and disability<br>• Minority status and language<br>• Housing type and transportation<br>• Epidemiologic factors<br>• Healthcare system factors | Constant feature | Surgo Foundation |
| Population Activities | Point-of-interest visits (POI Visits) | Time-dependent feature | SafeGraph |
| | Urban Activity Index (UAI)<br>• Work<br>• Social<br>• Home<br>• Traffic | Time-dependent feature | Mapbox |
| | Social Distancing Index (SDI) | Time-dependent feature | SafeGraph |
| | Venables Distance (VD) | Time-dependent feature | Mapbox |
| Human Mobility | Shelter-in-place Index (SIP) | Time-dependent feature | Spectus |
| | County Mobility Index (CMI) | Time-dependent feature | Spectus |
| | Colocation degree centrality (CDC) | Time-dependent feature | Meta |
| Social Interaction | Social Connectedness Index | Constant feature | Meta |

| | (SCI) | | |
|---|---|---|---|
| Disease Attribute | Reproduction Number (R0) | Time-dependent feature | U.S. Centers for Disease Control and Prevention |

### 3.2.1 Features related to social demographics

**(1) Population Density (PD)**

Population density of each county is calculated by dividing population square miles. Previous works have proven that PD is an important factor influencing the spread of an epidemic (Rocklöv and Sjödin 2020). PD was calculated based on the county-level Social Vulnerability Index of 2018 published by the United States Centers for Disease Control and Prevention (U.S. CDC) (Prevention 2020).

**(2) Gross Domestic Product (GDP)**

Previous works have shown that GDP could be a vulnerability index for COVID-19. Counties with higher GDP usually have a more robust economy and better health systems compared with counties with lower GDP (Sarmadi, Marufi et al. 2020). We used the 2018 county-level GDP published by the U.S. Department of Commerce (Commerce 2018).

**(3) Overall COVID-19 Community Vulnerability Index (CCVI)**

This study incorporates the county-level CCVI developed by Surgo Foundation based on CDC data (Foundation 2020), which comprises six social and demographic features determined by previous studies to affect the spread of COVID-19 with equal weights. CCVI is a composite score that reflects the extent of a county's vulnerability to COVID-19.

- **Socioeconomic status.** It is a feature accounting for a population's education, income, and occupation. Surgo Foundation developed this feature based on the CDC's Social Vulnerability Index, which accounts for populations below the poverty line, unemployed, and without a high school diploma.
- **Household composition and disability.** Surgo Foundation developed this feature based on the CDC's Social Vulnerability Index. This feature accounts for populations aged 65 or older, populations aged 17 or younger, populations older than 5 years of age with a disability, and single-parent households.
- **Minority status and language.** This feature accounts for minority and populations who speak English less than well based on CDC's Social Vulnerability Index.
- **Housing type and transportation.** Based on CDC's Social Vulnerability Index, this feature accounts for the population's housing types, such as multi-unit structures, mobile homes, and crowded housing. It also accounts for populations without vehicles and those who live in group quarters.
- **Epidemiologic factors.** Developed by Surgo Foundation in response to COVID-19, this feature accounts for populations with underlying conditions (e.g., cardiovascular, respiratory, immunocompromised, obesity, and diabetes) that are vulnerable to COVID-19.
- **Healthcare system factors.** Developed by Surgo Foundation for COVID-19, this factor accounts for poor health system capacity, strength, and preparedness.

### 3.2.2 Features related to population activities

**(1) Points-of-interest visits**

POI data provided by SafeGraph was used to examine population visits to POIs. SafeGraph aggregates POI data from diverse sources (e.g., third-party data partners, such as mobile application developers), and removes private identity information to anonymize the data. The POI data included basic information of a POI, such as the location name, address, latitude, longitude, brand, and business category. In this paper, we used the total number of visits by week to each POI in Weekly Pattern Version 2 to examine population visits to POI across 2787 counties in the United States (SafeGraph 2020). Furthermore, to remove the influence of disparate numbers of POIs in each county, we used the percentage change based on baseline POI visits of the first week, the week of March 3, 2020.

**(2) Urban activity index**
- Mapbox data was used to calculate the UAI. Mapbox data provides contact activity metrics in pre-defined tiles (measuring about 100 by 100 meters square) in 4-hour temporal resolution (Gao, Fan et al. 2021). We classified tiles into four categories. We then calculated an aggregated contact activity metric in those tiles to reveal four urban activities on a larger scale: social activity, traffic activity, home activity, and work activity. **Social tiles.** We classified tiles as social tiles in areas where at least one POI in SafeGraph is located.
- **Traffic tiles.** Traffic tiles includes tiles incorporating roads.
- **Home tiles.** Home tiles include residual buildings or have device information from 7 p.m. to 3 a.m.
- **Work tiles.** Work tiles show no activity in the evening.

**(3) Social distancing index**
Social distancing metrics developed by SafeGraph were used to calculate the SDI of each county (SafeGraph 2020). The SDI was calculated by dividing the number of cell phones within a household by the total number of devices within a county. Also, we used the percentage change based on the SDI of the first week to remove the potential influence of disparate numbers of devices in each county.

**(4) Venables distance (VD)**
Mapbox data was used to calculate the daily VD of each county according to Equation 1, reflecting the concentration of population activities(Louail, Lenormand et al. 2015).

$$D_V(t) = \frac{\sum_{i<j} s_i(t)s_j(t)d_{ij}}{\sum_{i<j} s_i(t)s_j(t)} \quad (1)$$

where $s_i(t)$ and $s_j(t)$ are the daily average activity intensities in cells $i$ and $j$, respectively, and $d_{ij}$ is the distance between the two cells. The resolution of the cell is 4 square meters. We also calculated the percentage change based on the values of the first week to remove the influence of disparate tiles and cells in each county.

### 3.2.3 Features related to human mobility

**(1) Shelter-in-place index**
SIP provided by Spectus was used as the feature of population mobility within counties (Cuebiq 2020). Spectus calculated the percentage of users in each county who traveled less than 330 feet as the SIP.

**(2) County mobility index**
CMI provided by Spectus was used as the feature of population mobility within counties (Cuebiq 2020). The CMI of each county is the median of aggregated movements of each user in a day in the county. For example, a CMI of 5 for a county represents that the median user in that county travels $10^5$ meters (100 kilometers).

**(3) Colocation degree centrality**
Facebook county-level colocation maps were used to calculate the colocation degree centrality

(CDC) of each county (Fan, Lee et al. 2021). Colocation maps could represent a network in which nodes are counties and edges represent the probability of population contacts between two counties. The network is undirected, and we calculated weighted colocation degree centrality as the feature reflecting mobility across counties. Also, we calculated the percentage change of the colocation degree centrality with respect to the first week.

### 3.2.4 Features related to social interaction

The county-level social connectedness index (SCI) provided by Facebook is used to account for social network structures affecting epidemic transmissions (Facebook 2020). The SCI for two counties is calculated according to Equation 2.

$$SCI_{i,j} = \frac{FB\_Connections_{i,j}}{FB\_Users_i \times FB\_Users_j} \tag{2}$$

We can find from Equation 2 that SCI of counties $i$ ($FB\_Users_i$) and $j$ ($FB\_Users_j$) is determined based on the number of Facebook connections (i.e., friends in Facebook) between two counties divided by the number of Facebook users in two counties. The SCI, therefore, reflects the strength of social connection between two counties. Then we mapped a fully connected network based on the SCI. The nodes in the network are counties, and edge weights are SCIs between counties. Finally, we calculated the weighted degree centrality of each county as the SCI feature.

### 3.2.5 Features related to disease attribute

The reproduction number ($R_0$) is an attribute of infectious diseases which estimates the number of secondary cases infected by the first case (Dietz 1993). We calculated the reproduction number of COVID-19 according to Equation 3 based on a simple epidemic transmission model (Ramchandani, Fan et al. 2020). The model assumes that one case would infect $R_0$ cases after a time interval $\tau$. Then $i(0)$ infected cases at the time step 0 will lead to $i(t) = i(0)R^{t/\tau}$ number of infected cases at time step t.

$$R_0 = e^{K\tau} \tag{3}$$

where $K = (\ln i(t) - \ln i(0))/t$, and we used $i = 5.1$ days for COVID-19 (Zhang, Litvinova et al. 2020). Furthermore, we used the percentage changes of $R_0$ with respect to the first week as the feature inputted in the model.

## 4. Methods and Model Specification

The attribute network embedding model created in this study has two main components: (1) county-level features (node features); and (2) cross-county movement network (spatial network topology). Figure 1 illustrates the components of the model. The attribute matrix was obtained in the first step and was calculated into the attribute similarity matrix. Second, a spatial network structure was constructed based on cross-county movements and was then expressed as an adjacency matrix. Third, the Accelerated Attributed Network Embedding (AANE) algorithm was adopted to capture the typological features as well as node attributes. Next, clustering analysis was performed on the attributed network embeddings every week. During this period, clustering results were computed, and counties with the same clustering tendencies were merged into archetypes. Finally, Kruskal Wallis and Dunn's tests were conducted to explore the archetype differences and expose features underlying distinct transmission risk patterns across archetypes.

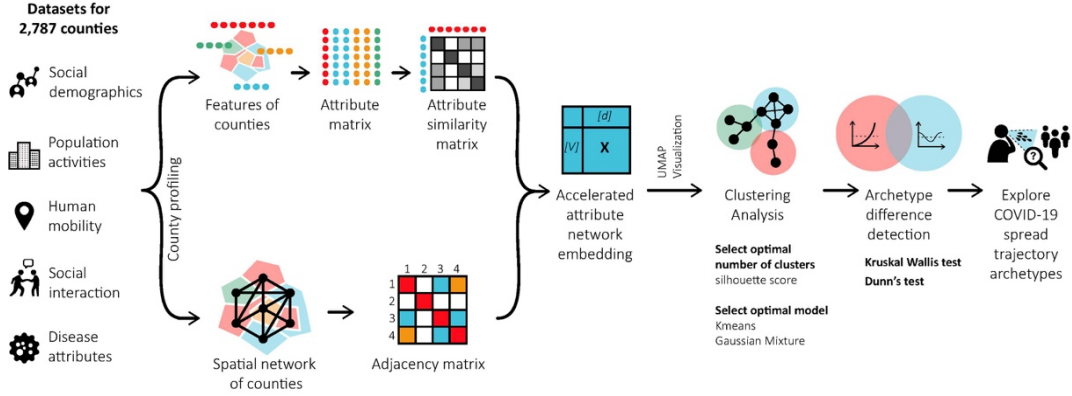

**Fig. 1.** Illustration of analytical framework.

**4.1 Constructing cross-county movement network**

In this step, a cross-county movement network was generated as the spatial network in the attributed network embedding model. The location and movement information were obtained from SafeGraph, Inc., a location intelligence data company that builds and maintains accurate point-of-interest data and store locations for the United States. The dataset contains POI attributes, including POI IDs, location names, and addresses. All POIs are labeled with Federal Information Processing System (FIPS) Codes for States and Counties, which uniquely identify geographic areas. In this paper, we used information of weekly POI visits across 2787 counties from March 3 to June 29, 2020. in the United States. The movements were aggregated at the county level to construct an undirected and weighted network that represented the sum of visit flows among counties. In this network, nodes are the centroid of each county and links are visitations between counties for one week. The weights of links capture the number of visits. The network is defined as

$$G = (V, E, W) \qquad (4)$$

where $V$ represents all of the counties and $E$ represents all the population flows between pairs of counties. The edge weights $W$ correspond to the counts of flows between two counties. For each individual trip on edge, the weight is incremented by 1.

**4.2 Accelerated Attributed Network Embedding**

Network embedding maps the topological structure of each node into a low-dimensional vector representation, preserving the original network proximity (Yan, Xu et al. 2006). It has been shown that network embedding could benefit various tasks, such as node classification (Zhou, Huang et al. 2006, Tang, Aggarwal et al. 2016) and network clustering (Narayanan, Belkin et al. 2006, Von Luxburg 2007). Given the characteristics of the cross-county movement network, which is large-scale with a large number of nodes and high-dimensional features, we adopted the Accelerated Attributed Network Embedding algorithm to extract the low-dimensional representation of the network and county-level disease-spread features.

The AANE algorithm, which is potentially helpful in learning a better embedding representation, was developed by (Huang, Li et al. 2017) to enable joint learning process to get attributed embeddings. The basic idea of AANE is to represent nodes as continuous vectors based on the decomposition of attribute affinity matrix and the penalty of embedding difference between connected nodes (Huang, Li et al. 2017).

Let $G = (V, E, W)$ be a network, where $V$ and $E$ are sets of nodes and edges, and $\omega_{ij} \epsilon W$ corresponds to edge $e_{ij}$ and reflects similarity between two nodes. Motivated by orthogonal similarity diagonalization theory of real symmetric matrixes, the AANE algorithm decomposes the semi-definite symmetric matrix $A$ into the following form: $A=H\Lambda H^T$, where $H$ is an orthogonal matrix and $\Lambda$ is a diagonal matrix (Yang and Lei 2021). Furthermore, AANE defines a new matrix $B$, whose elements are the square roots of the elements in *Lambda*, then:

$$A = H\Lambda H^T = HB^2H^T = HBH^T HBH^T = (HBH^T)(HBH^T)^T = UU^T \quad (5)$$

AANE uses cosine function to calculate the similarity matrix of nodes $S$, which is a semi-definite symmetric matrix (Huang, Li et al. 2017). Based on the above analysis, we can have:

$$S = QQ^T \quad (6)$$

Also, AANE considers that nodes with more similar topological structures or those connected by higher weights are more likely to have similar vector representations (Huang, Li et al. 2017). Thus, the objective function corresponding to the optimization problem is given below:

$$L = \|S - QQ^T\|_F^2 + \lambda \sum \omega_{ij} \|q_i - q_j\|^2 \quad (7)$$

where $\lambda$ is a balance parameter. Because the objective function has a specially designed structure that enables it to be optimized in an efficient and distributed manner, AANE adds a copy $Z=Q$ and the above function can be changed as follows (Yang and Lei 2021):

$$L = \sum \|S_i - q_i Z\|_2^2 + \lambda \sum \omega_{ij}\|q_i - z_i\|^2 + \frac{\rho}{2}\sum (\|q_i - z_i + u_i\|_2^2 - \|u_i\|_2^2) \quad (8)$$

where $\lambda$ is a balance parameter, $\rho$ is a penalty parameter, and $u_i \epsilon R^l (i = 1,2,...,n)$ is the scaled dual variable. The alternating directions method of multipliers is used to solve this problem (Yang and Lei 2021). By taking the derivatives of $q_i$ and $z_i$, the iterative formulas can be obtained as follows:

$$q_i^{t+1} = \left(2s_i Z^t + \lambda \sum \frac{\omega_{ij} z_j^t}{\|q_i^t - z_j^t\|_2}\right) + \rho(z_i^t - u_i^t)P^{-1} \quad (9)$$

$$P = 2(Z^t)^T Z^t + (\lambda \frac{\omega_{ij}}{\|q_i^t - z_j^t\|_2 + \rho})I \quad (10)$$

$$z_i^{t+1} = \left(2s_i Q^{t+1} + \lambda \sum \frac{\omega_{ij} q_j^{t+1}}{\|z_i^t - q_j^{t+1}\|_2}\right) + \rho(q_i^{t+1} + u_i^t)L^{-1} \quad (11)$$

$$L = 2(Q^{t+1})^T Q^{t+1} + (\lambda \frac{\omega_{ij} q_j^{t+1}}{\|z_i^t - q_j^{t+1}\|_2 + \rho})I \quad (12)$$

where $A^T$ means the transpose of $A$, $A^{-1}$ means the inverse of $A$ and $t$ means the $t$th iteration.

In this study, AANE embeds the adjacency matrix (cross-county human visitation network) and attributed similarity matrix (county-level standardized features) jointly for county's COVID-19 spread risk representation with the setting of 256 dimensions of embeddings.

### 4.3 Clustering analysis
#### 4.3.1 Optimal number of clusters

After implementing AANE embedding, we need to determine the level of similarity among counties or called optimal number of clusters. In this study, we performed two clustering methods, K-means and Gaussian mixture, to divide county-level COVID-19 spread risk patterns into clusters where inter-cluster similarities are minimized while the intra-cluster similarities are maximized.

K-means is an iterative method which minimizes the within-class sum of squares for a given number of clusters (Hartigan and Wong 1979). The algorithm starts with an initial guess for $K$ cluster centers $(u_1, u_2, u_3, \ldots \ldots, u_k)$. Then, each observation is placed in the cluster to which it is closest, i.e., find the mass center with the closest Euclidean distance for each cluster center.

$$label_i = \mathop{\arg\min}_{1 \leq j \leq k} \|x_i - \mu_j\| \tag{13}$$

Next, the cluster centers are updated as the average in the cluster.

$$\mu_j = \frac{1}{|c_j|} \sum_{i \epsilon c_j} x_i \tag{14}$$

Then, the above steps are repeated until the cluster centers no longer move.

As one of the most commonly used clustering algorithms, K-means has the features of simplicity, good understanding, and fast operation speed; however, the initial $K$ value has to be specified manually at the beginning.

Gaussian mixture model (GMM) is a probabilistic clustering method that calculates the probability that $n$ points are softly assigned to $K$ clusters. GMM assumes that all data points are generated by combining $k$ mixed multivariate Gaussian distributions into a mixture distribution.

$$p(x) = \sum_{i=1}^{k} \alpha_i \cdot p(x|\mu_i, \Sigma_i) \tag{15}$$

where $p(x|\mu_i, \Sigma_i)$ is the probability density function of a *n*-dimensional random vector $x$ that follows a Gaussian distribution.

$$p(x) = \frac{1}{(2\pi)^{\frac{n}{2}}|\Sigma|^{\frac{1}{2}}} e^{-\frac{1}{2}(x-\mu)^T \Sigma^{-1}(x-\mu)} \tag{16}$$

where $\mu$ is a *n*-dimensional mean vector, $\Sigma$ is $n \times n$ covariance matrix.

Therefore, $\mu_i$ and $\Sigma_i$ in equation (15) are the parameters of the *i*th Gaussian mixture component, and $\alpha_i > 0$ is the corresponding mixture coefficient.

$$\sum_{i=1}^{k} \alpha_i = 1 \tag{17}$$

Then, the process of GMM is to derive the parameters of each mixture component (i.e., the mean vector $\mu$, the covariance matrix $\Sigma$, and the mixture coefficient $\alpha$) by a certain parameter estimation method for a pre-determined the number of clusters $K$. Each multivariate Gaussian distribution component corresponds to one of the clusters.

According to (Brock, Pihur et al. 2008), it is rare in practice that the number of clusters $K$ is known at the beginning of the experiments. One possibility of identifying the most suitable number of clusters is the average silhouette method. The method calculates how well each object lies within its cluster using the index of silhouette score. The silhouette score measures the degree of confidence in the clustering assignment of a particular observation $i$, with well-clustered observations having values near 1 and poorly clustered observations having values near $-1$. The optimal number of clusters $K$ is the one that maximizes the average silhouette score over a range of possible values for $K$. For observation $i$, it is defined as

$$S(i) = \frac{b_i - a_i}{\max(b_i, a_i)} \tag{18}$$

where $a_i$ is the average distance between $i$ and all other observations in the same cluster, and $b_i$ is the average distance between $i$ and the observations in the nearest neighboring cluster, i.e.

$$b_i = \min_{C_k \in \iota \backslash C(i)} \sum_{j \in C_k} \frac{dist(i,j)}{n(C_k)} \tag{19}$$

where $C(i)$ is the cluster containing observation $i$, $dist(i, j)$ is the distance between observations $i$ and $j$, and $n(C)$ is the cardinality of cluster $C$.

In this study, we used K-means and Gaussian mixture to perform clustering test on the weekly datasets. Finally, the best $K$ value represented by the maximum silhouette score, which was averaged over all weeks, was selected as the optimal number of clusters.

### 4.3.2 Optimal clustering method

Further, we need to determine the optimal clustering model for underlying the optimal clustering pattern of county-level COVID-19 spread risk.

A variety of stability measures, including average proportion of non-overlap (APN), average distance (AD), average distance between means (ADM), and figure of merit (FOM), aiming at validating the results of a clustering analysis and determining which algorithm performs best have been proposed by (Von Luxburg 2010). The stability measures compare the results from clustering based on the full data to clustering based on removing each column, one at a time. These measures work especially well if the data are highly correlated, which is often the case in high-resolution human mobility data. In all cases, the average is taken over all the deleted columns, and all measures should be minimized (Datta and Datta 2003).

**Average proportion of non-overlap (APN)**

The APN measures the average proportion of observations not placed in the same cluster by clustering based on the full data and clustering based on the data with a single column removed. Let $C^{i,0}$ represents the cluster containing observation $i$ using the original clustering and $C^{i,l}$ represents the cluster containing observation $i$ where the clustering is based on the dataset with column $l$ removed. Then, APN is defined as

$$APN(K) = \frac{1}{MN} \sum_{i=1}^{N} \sum_{l=1}^{M} (1 - \frac{n(C^{i,l} \cap C^{i,0})}{n(C^{i,0})}) \qquad (20)$$

The APN is in the interval [0, 1] with values close to 0 corresponding with highly consistent clustering results.

**Average distance (AD)**

The AD computes the average distance between observations placed in the same cluster by clustering based on the full data and clustering based on the data with a single column removed. It is defined as

$$AD(K) = \frac{1}{MN} \sum_{i=1}^{N} \sum_{l=1}^{M} \frac{1}{n(C^{i,0})n(C^{i,l})} \left[ \sum_{i \in C^{i,0}, j \in C^{i,j}} dist(i,j) \right] \qquad (21)$$

The AD has a value between 0 and ∞ and smaller values are preferred.

**Average distance between means (ADM)**

The ADM computes the average distance between cluster centers for observations placed in the same cluster by clustering based on the full data and clustering based on the data with a single column removed. It is defined as

$$ADM(K) = \frac{1}{MN} \sum_{i=1}^{N} \sum_{l=1}^{M} dist(\bar{x}_{C^{i,l}}, \bar{x}_{C^{i,0}}) \qquad (22)$$

where $\bar{x}_{C^{i,0}}$ is the mean of the observations in the cluster which contain observation $i$ and $\bar{x}_{C^{i,l}}$ is similarly defined. It also has a value between 0 and ∞ and smaller values are preferred.

**Figure of merit (FOM)**

The FOM measures the average intra-cluster variance of the observations in the deleted column, where the clustering is based on the remaining samples. This estimates the mean error using predictions based on the cluster averages. For a particular left-out column $l$, the FOM is

$$FOM(l, K) = \sqrt{\frac{1}{N} \sum_{k=1}^{K} \sum_{i \in C_k(l)} dist(x_{i,l}, \bar{x}_{C_k(l)})} \qquad (23)$$

where $x_{i,l}$ is the value of the $i$-th observation in the $l$th column in cluster $C_k(l)$ and $\bar{x}_{C_k(l)}$ is the average of cluster $C_k(l)$. FOM has a value between 0 and ∞ with smaller values equaling better performance.

### 4.4 Archetype difference detection

Kruskal Wallis and Dunn's tests were performed each week based on weekly datasets from March 3 to June 29, 2020 (17 weeks in total). Each county would be assigned to a specific cluster each week and thus assembled into a temporal distribution of clusters over the 17 weeks. Next, counties with same temporal distributed clusters were merged into archetypes to explore dynamical COVID-19 spread risks over the entire period. Then, we used Kruskal Wallis test and Dunn's tests to explore the differences of spread risks among archetypes combined from clusters. The two methods were used sequentially, in which Kruskal Wallis test was used to explore whether the features statistically differ among the archetypes, while Dunn's test was used to detect in which archetypes the features had significant differences.

#### 4.4.1 Kruskal Wallis (KW) test

The KW test explores the null hypothesis that the population median of all of the groups is equal, which is a non-parametric alternative to the one-way ANOVA test when we have two or more independent groups (Kruskal and Wallis 1952). As the *p*-value is less than the significance level 0.05, we can conclude that there are significant differences between the clusters. In KW test, all the data are pooled and ranked from smallest (1) to largest ($N$), then the sums of ranks in each subgroup are added up, and the probability is calculated. The statistic $H$ is

$$H = \frac{12}{N(N+1)} \sum \frac{R^2_i}{n_i} - 3(N+1) \qquad (24)$$

where $N$ is the total number, $n_i$ is the number in the *i*-th group, and $R_i$ is the total sum of ranks in the *i*-th group.

#### 4.4.2 Dunn's test

Once KW test finds a significant difference among two or more groups, the Dunn's test can be used to pinpoint significant features. Dunn's test is a post hoc non-parametric test (i.e., it is run after an KW test). The Dunn index is the ratio of the smallest distance between observations not in the same cluster to the largest intra-cluster distance (Brock, Pihur et al. 2008), which is computed as

$$D(c) = \frac{\min_{C_k, C_l \in C, C_k \neq C_l} (\min_{i \in C_k, j \in C_l} dist(i,j))}{\max_{C_m \in C} diam(C_m)} \qquad (25)$$

where $diam(C_m))$ is the maximum distance between observations in cluster $C_m$. The Dunn index has a value between 0 and ∞ and should be maximized.

### 5. Results

#### 5.1 Clustering analysis

Since cross-county movement network had temporal variations, we constructed the networks for each week from March 3 to June 29, 2020. The nodes of each network are counties in the United States, and the edges represent the visits between counties. Since a higher volume of visits represents closer connection between the counties, the number of visits between counties are used to calculate the weights of the edge. County-level features were regarded as node attributes. Then, the AANE algorithm was applied to all weekly networks. Combining both topological structure and node attributes, this algorithm represented each node in the network as a 256-dimension vector. UMAP was used to reduce the dimension of the attributed embeddings as a two-dimension vector for visualization purpose.

This study performed cluster analysis to the nodes of weekly networks to classify counties based on their pandemic spread risk patterns. K-means and Gaussian mixture models were selected to cluster the nodes. To decide the most optimal number of clusters, we examined clustering the nodes into 2, 3, 4, 5, and 6 clusters using the two methods. Silhouette score was calculated to show the best number of clusters for both methods. Since the cluster analysis was performed on the nodes of weekly human movement network, this study calculated the average of silhouette scores during these weeks. Table 2 presents the number of clusters, cluster methods and the corresponding average silhouette scores. It can be seen that when clustering the nodes into two clusters using K-means method, highest average silhouette score is achieved as 0.4810. However, the silhouette score is not the only criterion when deciding on the cluster number. Clustering the nodes in two clusters could be too broad and lack the require interpretability to inform about differences in pandemic risk trajectories needed for policy formulation. Accordingly, we chose the number of clusters as four, which gives the second highest average silhouette score (0.4682) when using K-means method and the highest score (0.4613) when using Gaussian mixture method. Compared to two clusters, four clusters could reduce the information loss caused by over-generalization, which helps maintain more elaborated details. The decision to use four clusters was therefore based on consideration of both validity measures and the context of the study.

**Table 2**
Statistics of average silhouette scores of the two clustering methods.

| Number of clusters | Average silhouette score of K-means | Average silhouette score of Gaussian mixture |
|---|---|---|
| 2 | 0.4810 | 0.4569 |
| 3 | 0.4420 | 0.4341 |
| 4 | 0.4682 | 0.4613 |
| 5 | 0.4312 | 0.4137 |
| 6 | 0.4123 | 0.3932 |

When setting the number of clusters to four, the average silhouette score of K-means (0.4682) is slightly higher than that of Gaussian mixture method (0.4613). To further determine the cluster method, a stability test was performed. This study used four measures, including APN, AD, ADM, and FOM, for each week and then calculated the averages. The results demonstrated that K-means outperformed Gaussian mixtures, as can be seen in Table 3. Thus, clusters produced by K-means method were selected for further analysis.

**Table 3**
Statistics of stability measures of the two clustering methods.

| Measures | Average | Average AD | Average ADM | Average |
|---|---|---|---|---|

|  | APN |  |  | FOM |
|---|---|---|---|---|
| K-means | 0.4590 | 3.3216 | 2.1179 | 2.3863 |
| Gaussian mixtures | 0.4658 | 3.5093 | 2.1979 | 2.3929 |

The visualizations of the clusters are shown in Figure 2, from which one can see that the clusters are well separated. We also specified the representative county (Brazos County in Texas) to describe the features and pandemic risk trajectories within each cluster. Brazos County located in the cluster orange in the first four weeks (from March 3 to March 30), and then it moves to cluster red in the following two weeks (from March 31 to April 13) and cluster green in the seventh week (from April 14 to April 20). Next, Brazos County moves back to the cluster orange until the last week (from April 2 to June 29) in this study. It can be observed that even the same county may locate in different clusters during different weeks as the pandemic unfolds. The temporal variations in the clustering of counties are due to response to non-pharmaceutical policies and the spike in the number of cases as the disease spread during the first wave.

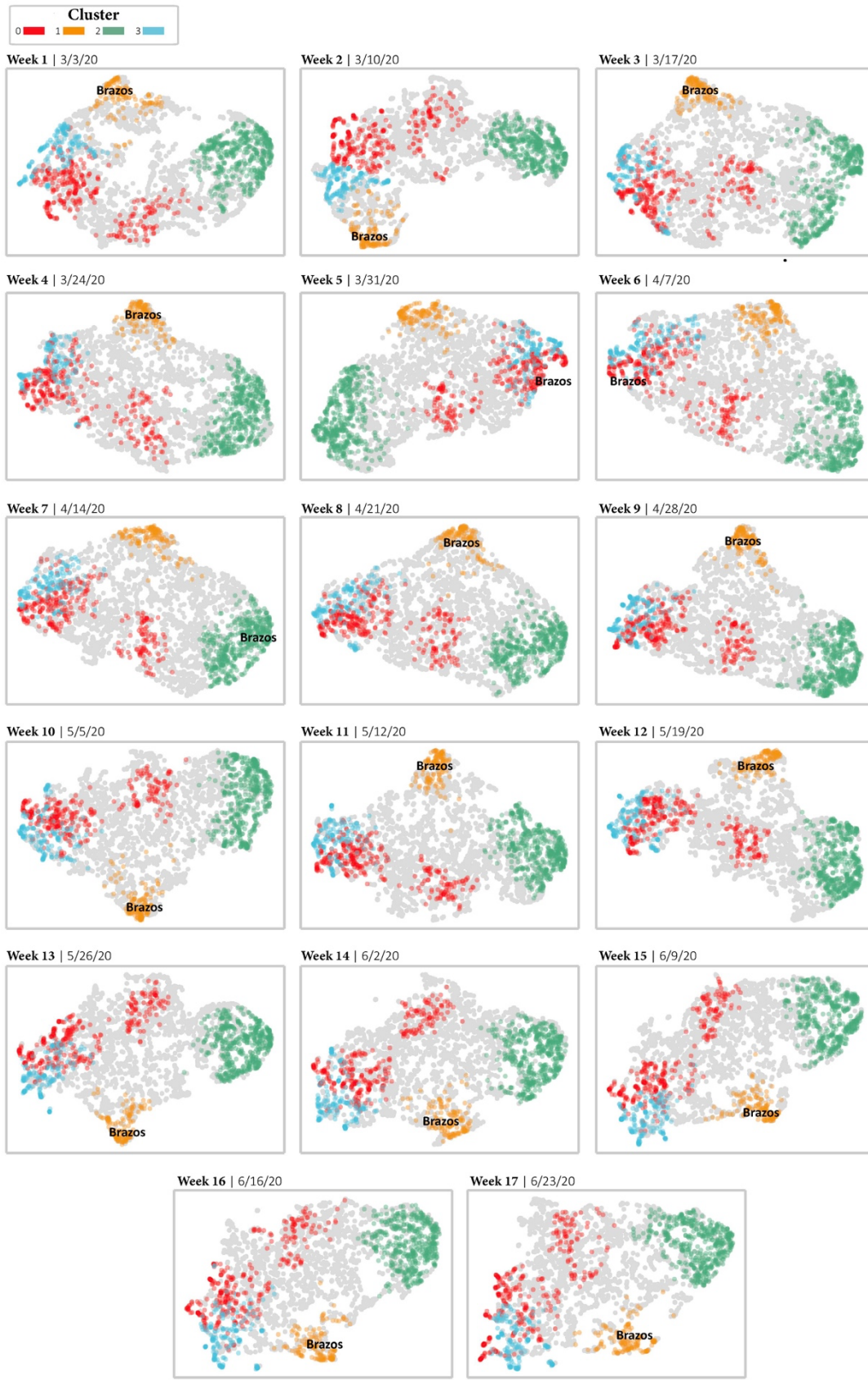

**Fig. 2**.
Visualization of the 4 clusters for 17 weeks from March 3 to June 29, 2020. Brazos

County in Texas located in different clusters during different weeks as the pandemic unfolds.

To take the temporal effects into consideration, we listed all clusters of counties during the 17-week period then merged the counties with the exact same tendencies into archetypes. Ten archetypes were identified from this step and the top 5 archetypes were shown in the left part of Figure 3. Then, the archetypes with less than 20 counties were removed to avoid potential contingence and increase result robustness. The remaining archetypes are shown on the right side of Figure 3. Since archetype 1 and archetype 4 have only slight differences, they were merged. Taking the archetype 0 as an example, it comprises 505 counties, which indicates that these counties have similar characteristics in terms of COVID-19 risk transmission dynamics over the 17-week time span. In the first three weeks (from March 3 to March 23), these counties located in the same cluster. Next, in the following 11 weeks (from March 24 to June 8), the 505 counties moved to another cluster. In the last three weeks (from June 9 to June 29), they returned to the original cluster. The counties which were included in the top 4 archetypes are shown in Figure 4.

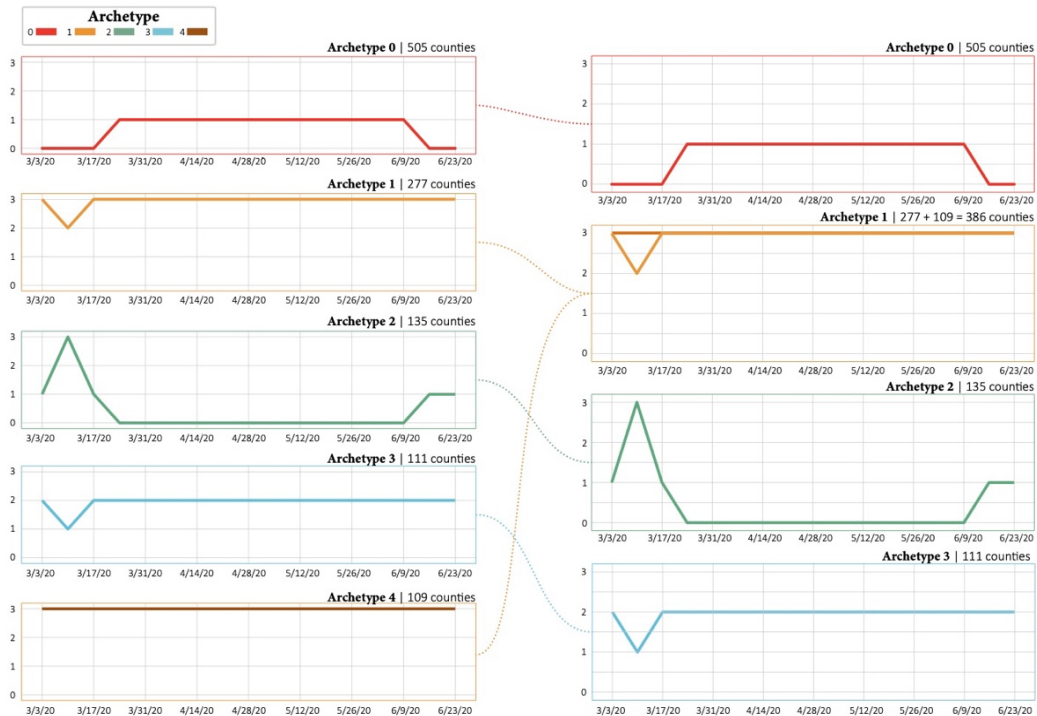

**Fig. 3.**
The temporal tendencies of county-level COVID-19 spread risk in archetypes. The left half is the initial archetypes and the right half is the merged archetypes.

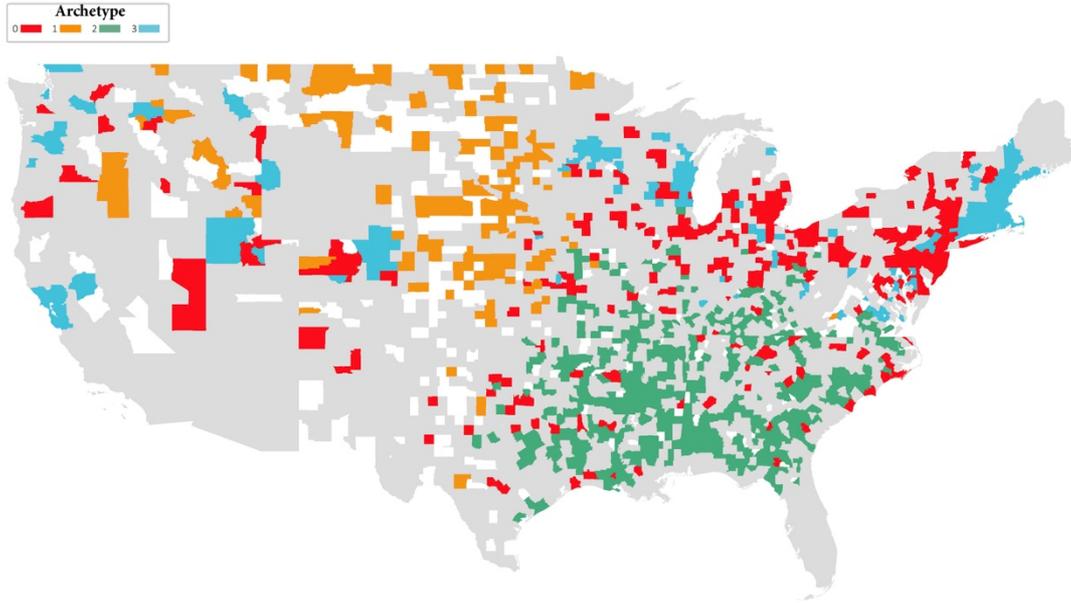

**Fig. 4.**
Spatial distribution of the counties that were included in the top 4 archetypes.

**5.2 Feature importance analysis**

Since each archetype represents a distinct pattern of COVID-19 transmission risk in communities, finding the most prominent features that differentiate archetypes from each other is essential. To this end, we adopted the Kruskal Wallis and Dunn's test to all the county-level features, as well as the number of weekly new infection cases recorded by the CDC. The Kruskal Wallis test was used to detect if features are statistically significant different among the archetypes. Table 4 listed the statistically non-significant results. It can be seen that only R0 and urban activity index (home) are not significantly different from other archetypes during some weeks, while for other features, the archetype differences of which all exist. The Kruskal Wallis test only detects the existence of archetype differences, while the Dunn's test could specify which archetypes features are different. Thus, to further examine the features which could be different across archetypes, the Dunn's test was performed.

**Table 4**
Features of similar distribution among clustered counties

| Week starting date | Features of similar distribution among clustered counties | Statistics | $p$-value |
|---|---|---|---|
| 2020-03-10 | R0 | 3.895 | 0.2731 |
| 2020-03-17 | R0 | 0.000 | 1.0000 |
|  | UAI home | 5.592 | 0.1333 |
| 2020-03-24 | R0 | 0.298 | 0.9604 |
| 2020-05-26 | UAI home | 4.360 | 0.2252 |
| 2020-06-02 | UAI home | 7.137 | 0.0677 |
| 2020-06-09 | UAI home | 6.330 | 0.0966 |
| 2020-06-16 | UAI home | 4.717 | 0.1937 |

Figure 5 shows the boxplot of new infected COVID-19 cases per 100k population among the four archetypes during the 17 weeks. In terms of new cases, the four archetypes show distinct characteristics that distinguish themselves from each other from both the growing tendency and deviation. During the first three weeks, all the archetypes had very few new cases; During the following six weeks, a rise of new cases per 100k population were observed from all the archetypes, while variations within each archetype were also observed. Archetype 0 and archetype 3 have smaller within-archetype variation, which means the new cases within counties falling in these two archetypes had a similar pattern. Yet for archetype 2 and archetype 3, the within-archetype variation turned to be more significant. The variation in the new cases is especially high in archetype 3. During the remaining eight weeks, the general patterns started to split. Archetypes 0 and 1 still observed fast growth of new infected cases, while archetypes 2 and 3 kept a rather low level of newly infected cases. From the perspective of within-group variation, archetype 2 and archetype 3 both have high variation, which means counties in these archetypes had less-consistent COVID spread patterns although they have similarities in features. Based on the two analysis dimensions, the results reveal four different COVID-19 spread patterns in communities: (1) counties with high and consistent increase, (2) counties with a high number of cases with fluctuating increase, (3) counties with mild and consistent increase, and (4) counties with mild and fluctuating increase.

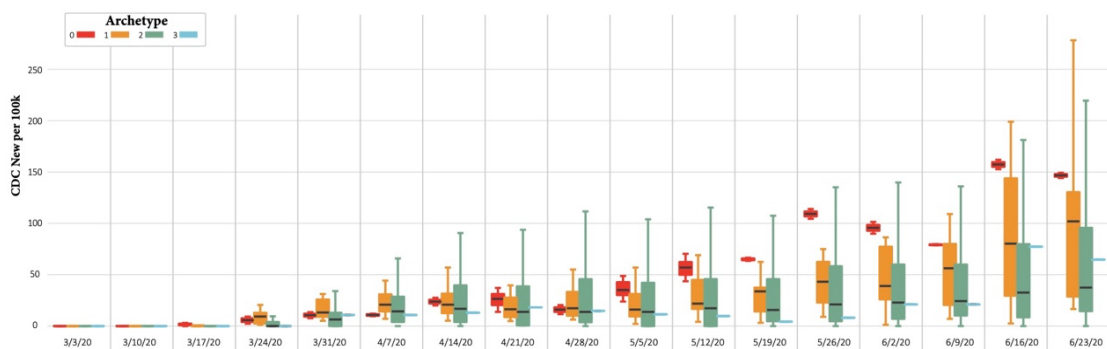

**Fig. 5.**
Distribution of new infected COVID-19 cases per 100k population among the four archetypes during the 17 weeks.

After identifying the spread pattern within each archetype, we performed a Kruskal Wallis test for all the county-level features and found only R0 and urban activity index (home) were not significantly different among the archetypes during several weeks, while for other features, differences in features across archetypes exist. Subsequently, Dunn's test was performed on the significantly different features to detect from which archetype features were different. Accordingly, we could find archetype differences in terms of their county-level features.

From the Dunn's test, we identified four features, including population density, gross domestic product (GDP), minority status and language, and POI visits, which exhibit significant differences across all the four archetypes. Since these features in different archetypes are distinct from each other, they can be used to best describe the characteristics of counties in each archetype. The high-increase archetype (marked as red and orange in Figures 6, 7, 8, and 9) is characterized as higher population density, higher GDP, greater number of POI visits, and larger percentage of minorities. This archetype showed a typical transmission risk pattern: highly developed economic centers with dense population increased the chance of interpersonal contact in such areas, which increased the number of new cases. The higher number of new cases in turn worsened the transmission trend. From the perspective of human movement activities, the percentage change of number of POI visits in red and orange archetypes is apparently higher than the other two archetypes, indicating that POI visits can be an important risk factor when considering COVID-19 spread. This observation is consistent with the insights of (Fan, Lee et al. 2021) and(Shi, Pain et al. 2022). Counties in archetype 2 and archetype 3 are characterized as mild increase areas, and that may relate to the lack of high-risk factor. Population in these counties are not as dense as those in archetype 0 and archetype 1, and lower GDP may indicate less economic activity, which reduced the spread chances of COVID-19 via interpersonal contact. Since studies have confirmed that points of interest can be hotspots of disease transmission (Jia, Lu et al. 2020), and super-spreaders of the virus (Zhou, Xu et al. 2020) are more likely to present at some of the POIs, smaller number of POI visits may help avoid radical increase of new COVID-19 cases. Moreover, different socio-demographic groups are more balanced in archetype 2 and archetype 3, which contributes to the ability of populations to follow protective actions (such as staying at home) that would decrease the number of cases.

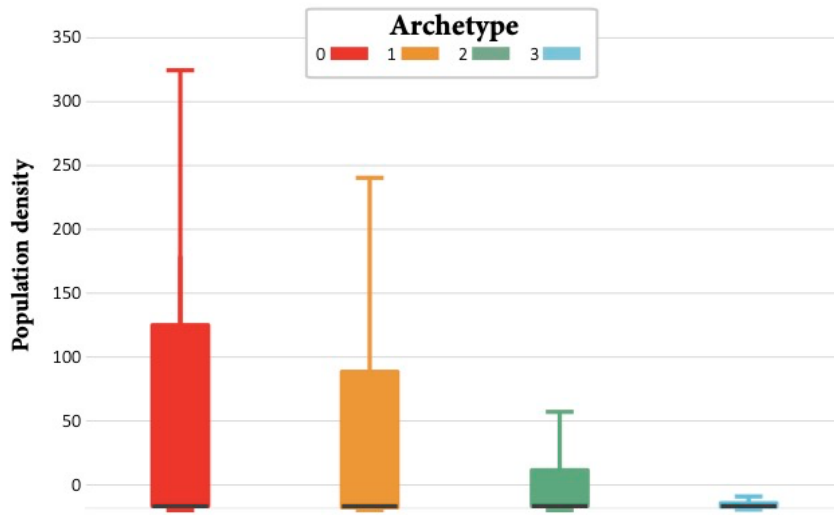

**Fig. 6.**
Distribution of population density across the four archetypes.

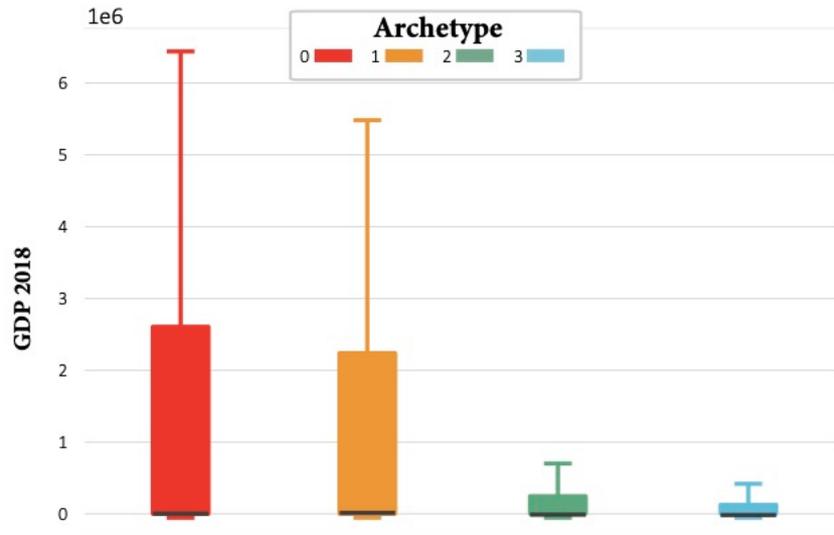

**Fig. 7.**
Distribution of gross domestic product across the four archetypes.

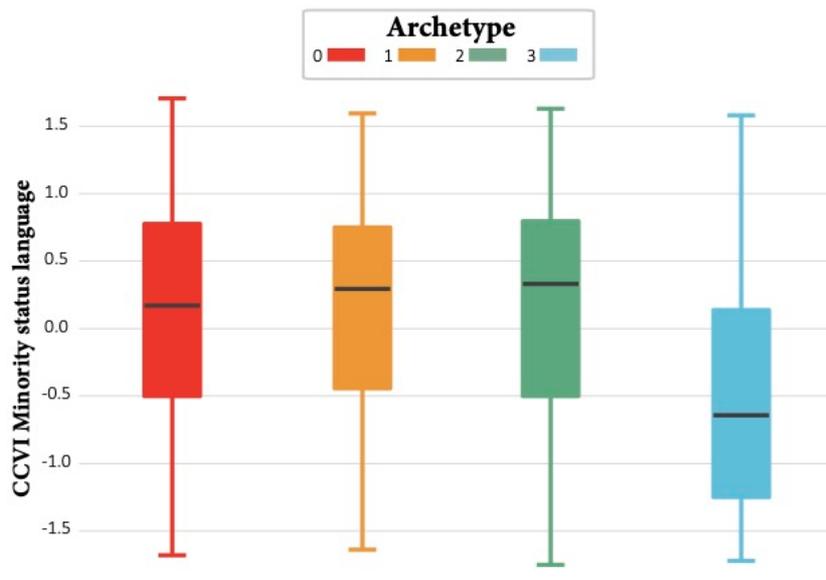

**Fig. 8.**
Distribution of minority status and language across the four archetypes.

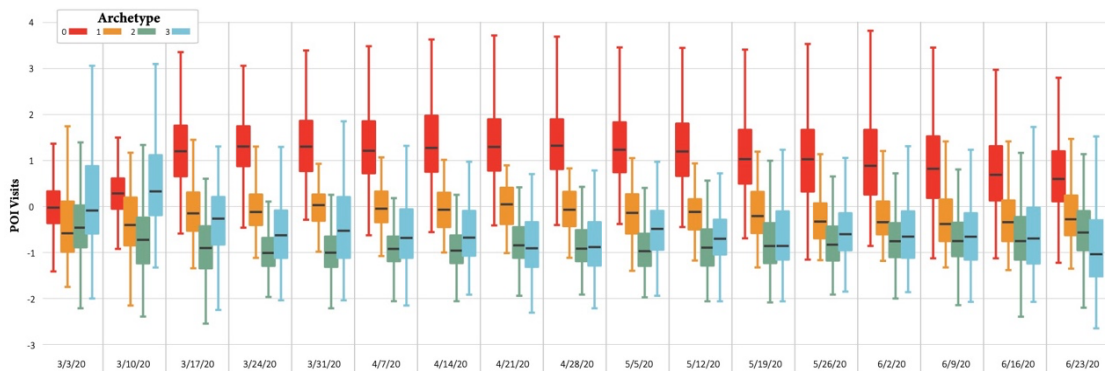

**Fig. 9.**
Distribution of POI visits across the four archetypes during the 17 weeks.

Further comparisons were made between the archetypes with similar growth tendency to examine the different features between the archetypes. Although the medians of newly infected cases in archetype 0 and archetype 1 are both high, archetype 1 has a much larger within-cluster variation, which means counties in this archetype had less consistent new case growth tendency. Archetype difference detection shows that the variation of healthcare system factors (one sub-feature in the overall COVID-19 Community Vulnerability Index) of counties in archetype 1 is significantly larger than that in archetype 0, which differentiates the two archetypes from each other. The results show that a high while inconsistent archetype also has varied healthcare system vulnerability level. This influence may be explained by the COVID-19 test capacity or the capacity of taking care of the infected and preventing larger scale infection. Counties without that capacity would have high but inconsistent patterns of new case growth. The feature that distinguishes archetype 2 and archetype 3, is the reproduction number ($R_0$). The percentage change of $R_0$ has more deviation in archetype 2 than that in archetype 3. In other words, the mild but inconsistent archetype has inconsistent $R_0$ change tendency.

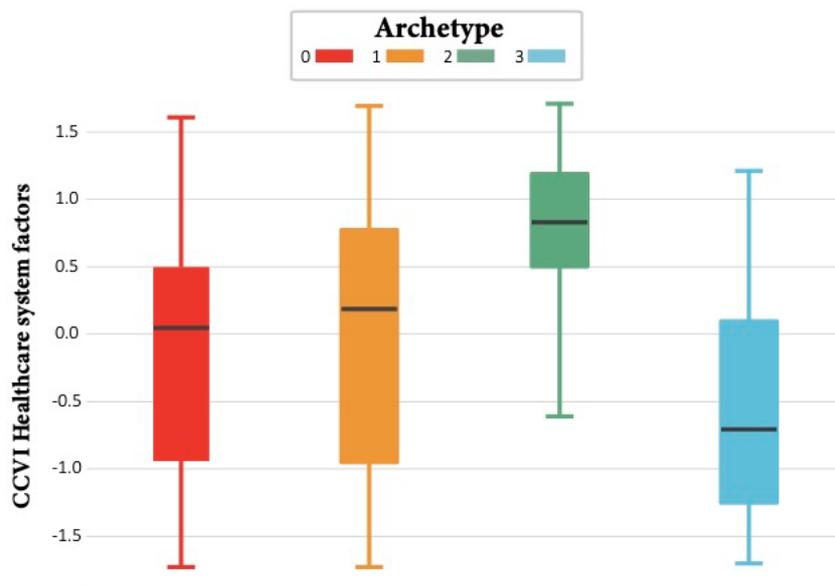

**Fig. 10.**
Distribution of healthcare system factors across the four archetypes.

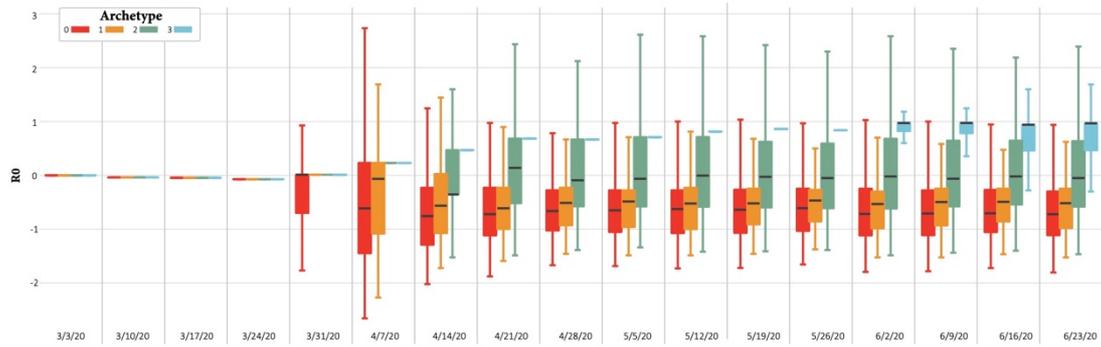

**Fig. 11.**
Distribution of reproduction number (R0) across the four archetypes during the 17 weeks.

## 6. Concluding remarks

Pandemic risk trajectories are not homogenous among all cities and communities. Various features could influence the differences in pandemic trajectories. The heterogeneity of pandemic spread trajectories motivated this study to explore different of COVID-19 spread trajectories in the 2787 counties in the United States during the first pandemic wave and also to uncover the county-level features that contribute distinctive pandemic spread trajectories across different clusters.

The study and findings have multiple important contributions: first, the findings expose four main COVID-19 spread trajectory patterns in the United States which signify the importance of recognizing the heterogeneity in pandemic risk trajectories of different areas for predictive pandemic monitoring and policy formulation. Formulating national one-size-fits-all policies for all counties which follow different spread trajectories would not yield the desired pandemic control outcomes. Public health officials, county health department officials, and policymakers can account for differences among different counties based on their heterogeneous features and formulate place-based policies consistent with population activity, cross-county movement, and socio-demographic and disease-related features. These features can be proactively monitored as trigger indicators for initiating or halting policies given the spread trajectory of the pandemic. For example, counties with high GDP, dense population, larger proportion of minorities, and more active population activities may be at risk of rapid increase of new cases. By adopting the methodological framework proposed in this study, policymakers could become aware of the transmission risk in a timely manner and then use the data as a reference for dynamically adjusting COVID-19 related public policies.

Second, the attributed network embedding model contributes to advancing artificial intelligence and machine-learning techniques for data-driven pandemic management. Attributed network embeddings were calculated to integrate both typological characteristics of the human visitation network and place-based features. The model captures several heterogeneous features related to population activity, mobility, sociodemographic, economic, and disease-related attributes and their non-linear interactions, as well as the cross-county movement network characteristics, all of which contribute to the spread trajectories of the pandemic. Such complex feature interactions cannot be fully captured using the existing models. Clustering analysis performed on the attributed network embeddings uncovered four archetypes of spread risk patterns, and four features—population density, gross domestic product, minority status and language, and number of POI visits—that contribute to distinctive transmission risk archetypes across the United States.

**Acknowledgement**

The authors would like to acknowledge funding support from the National Science Foundation RAPID project #2026814: Urban Resilience to Health Emergencies: Revealing Latent Epidemic Spread Risks from Population Activity Fluctuations and Collective Sense-making, and Microsoft AI for Health COVID-19 Grant for cloud-computing resources. The authors would also like to acknowledge that Spectus, Mapbox, and SafeGraph provided mobility and population activity data. Any opinions, findings, conclusions, or recommendations expressed in this material are those of the authors and do not necessarily reflect the views of the National Science Foundation, Microsoft, SafeGraph, Spectus, or Mapbox.

**Data availability**

The data that support the findings of this study are available from SafeGraph, Mapbox, and Spectus, but restrictions apply to the availability of these data, which were used under license for the current study. The data can be accessed upon request submitted to the data provides. Other data we use in this study are all publicly available.

**Code availability**

The code that supports the findings of this study is available from the corresponding author upon request.